\documentclass{article}

    \PassOptionsToPackage{numbers, compress}{natbib}
  \usepackage[preprint]{neurips_2026}

\usepackage[utf8]{inputenc} 
\usepackage[T1]{fontenc}    
\usepackage{hyperref}       
\usepackage{url}            
\usepackage{booktabs}       
\usepackage{amsfonts}       
\usepackage{nicefrac}       
\usepackage{microtype}      
\usepackage{xcolor}         
\usepackage{amsmath}
\usepackage{amssymb}
\usepackage{graphicx}
\usepackage{algorithm}
\usepackage{algpseudocode}

\title{DAGGER: Gradient-Free Construction of Transiently Amplifying Networks under Hard Connectivity Constraints}

\author{%
  James C.~Ferguson\thanks{The African Institute for Mathematical Sciences (Muizenberg, South Africa) and the Institute of Science and Technology Austria (Klosterneuburg, Austria)} \\
  \texttt{jamesf@aims.ac.za} \\
}

\begin{document}

\maketitle

\begin{abstract}
Many networks not only support but also rely on transient non-normal amplification, an orders-of-magnitude increase in the activity of an otherwise stable system.  Constructing such networks under hard sign/sparsity/diagonal constraints -- the regime relevant for biological connectomes and structured RNN initializations -- has so far required either gradient-based local search with thousands of inner-loop eigendecompositions or Schur-form direct construction in an abstract basis that breaks the constraints under projection.

Here we introduce DAGGER (Directed Acyclic Graph Guided Edge Reweighting), a gradient-free single-pass algorithm. Given a stable signed sparse matrix, DAGGER produces an output with the same sign, sparsity, and diagonal. A single scalar $\beta$ controls a Wasserstein-2 budget that smoothly trades exact multiset preservation ($\beta = 0$) for amplification; peak amplification grows essentially without bound with $\beta$, empirically reaching $10^{10}$ before numerical overflow.

DAGGER matches or exceeds gradient-based methods at multiset preservation in a single forward pass -- 30-100$\times$ fewer eigendecompositions than a typical gradient inner loop -- and at moderate $\beta$ beats them by orders of magnitude with connectivity exactly preserved. We develop the algorithm, compare it to the existing methods and on a downstream signal-detection task, and examine the diagnostics that show why DAGGER is structurally different from other amplifying networks.
\end{abstract}

\section{Introduction}
Many cortical phenomena depend on non-normal transient amplification, during which a brief input pulse drives a network's internal state to grow by orders of magnitude before stable decay returns the system to rest. This mechanism is believed to be important for a host of neuronal processes including  motor preparation in primate motor cortex \citep{hennequin2014, stroud2018}, OFF responses in auditory cortex \citep{bondanelli2021}, and balanced amplification in cortical sensory areas \citep{murphy2009}. On the machine-learning side, controlled non-normality has emerged as a tool for expanding RNN expressivity beyond orthogonal/unitary parameterizations while still preserving stability \citep{kerg2019,orhan2019}.
%

In each of these settings one faces a constrained design problem: given an initial matrix $A^0$ (typically stable, signed, and sparse), return a new matrix $A$ that is both stable and strongly amplifying yet ideally preserves as much of the original structure of $A^0$ as possible. Existing methods in this direction fall into three regimes: gradient-based local search \citep{hennequin2014, stroud2018}; Schur-form direct construction \citep{christodoulou2022,goldman2009}; and low-rank construction \citep{bondanelli2020}. However, none of these methods cleanly solves the constrained design problem.

Our contribution is: DAGGER (Directed Acyclic Graph Guided Edge Reweighting). DAGGER combines the constraint-respecting nature of gradient methods with the constructive simplicity of Schur methods, in a single forward pass with no iterative inner loop. 

We make four concrete contributions:
\begin{enumerate}
\item $\textbf{Algorithm:}$ DAGGER produces strongly amplifying networks in $O(n^3 \log(1/\varepsilon))$ time -- dominated by a single stability bisection -- compared to $O(n^3 \cdot \text{iters})$ for gradient methods. This corresponds to 30-100$\times$ fewer eigendecompositions than a typical gradient inner loop, translating to roughly 5$\times$ wall-clock speedup at $n = 300$ against our SOC variant with the gap widening under the canonical full-step setting.
\item $\textbf{Pareto knob:}$ A single scalar $\beta$ controls a Wasserstein-2 budget that smoothly trades multiset preservation for amplification. Setting $\beta = 0$ preserves the original magnitude multiset exactly; growing $\beta$ traces a Pareto front whose peak amplification grows essentially without bound (empirically reaching $10^{10}$ before float64 overflow).
\item $\textbf{Mechanistic topology preservation:}$ DAGGER's amplification gain comes from rearrangement, not from changing connectivity. Structural diagnostics show that at matched amplification, DAGGER networks have backward-edge weight fraction more than 40$\times$ lower than SOC networks -- confirming that DAGGER explicitly reshapes the connectivity towards feedforward structure within the fixed sparsity pattern.
\item $\textbf{Empirical validation:}$ DAGGER networks achieve a perfect area under the curve (AUC = 1, perfect discrimination, where 0.5 = chance) in a receiver operating characteristic (ROC) signal-detection task at signal amplitudes an order of magnitude weaker than other methods.
\end{enumerate}

We develop the algorithm and its empirical behaviour below: first we contextualize DAGGER among related methods; then we describe the algorithm; next we report experiments; and we conclude by discussing limitations and future directions.

\section{Related Work}

\subsection{SOC and gradient-based methods} 
The amplification problem we solve is most closely associated with the SOC framework introduced by \citep{hennequin2014} for cortical inhibitory stabilization, and extended by \citep{stroud2018} to motor primitive selection. SOC optimizes non-normal amplification within Dale's-law-constrained networks via projected gradient descent on a smoothed spectral abscissa \citep{vanbiervliet2009}, subject to sign constraints and element-wise magnitude bounds. The approach is constraint-respecting and standard in the computational neuroscience literature; its principal cost is the inner loop of potentially hundreds-to-thousands of eigendecompositions per construction. 

\subsection{Schur-form construction}
 The Schur-decomposition framework articulated in \citep{christodoulou2022} proposes independent manipulation of the eigenspectrum and the feedforward norm via $A = Q(D + T)Q^\top$, with $D$ block-diagonal carrying eigenvalues and $T$ strictly upper-triangular carrying the feedforward / non-normal structure. This is philosophically similar to DAGGER but operates in the orthogonal Schur basis rather than the original (signed sparse) basis, and the two are not interchangeable when constraints must be preserved exactly: projecting a Schur-amplified matrix onto a fixed sign/sparsity pattern destroys most of the amplification. \citep{goldman2009} gives the limiting case — a strictly upper-triangular nilpotent connectivity that integrates inputs without recurrent feedback — which represents a theoretical ceiling on the amplification achievable from a given magnitude multiset. \citep{bondanelli2020} builds minimal low-rank networks that produce targeted transient trajectories from input to readout — a complementary problem (targeted I/O design rather than generic substrate amplification). \citep{murphy2009} provides the architectural recipe of balanced amplification, a structural prescription rather than an optimization method.

\subsection{ML-side parameterizations and pseudospectral methods.} The non-normal RNN of \citep{kerg2019} parameterizes the recurrent weight matrix in Schur form for end-to-end backpropagation, with explicit normal/non-normal split; \citep{orhan2019} shows that sequential non-normal initialization improves long-range memory in RNNs. Orthogonal/unitary RNN families (\citep{arjovsky2016,mhammedi2017,lezcano2019}) take the opposite extreme -- fully normal connectivity -- and explicitly forbid amplification. Generic pseudospectral abscissa optimization (\citep{lewis1996,trefethen2005}) provides the abstract framework of which SOC is a specialized instance, typically not applied under hard sign and sparsity constraints. These methods address the training and analysis problems; DAGGER addresses the construction-from-substrate problem and is complementary.

\begin{figure}
  \centering
  \includegraphics[width=0.99\textwidth]{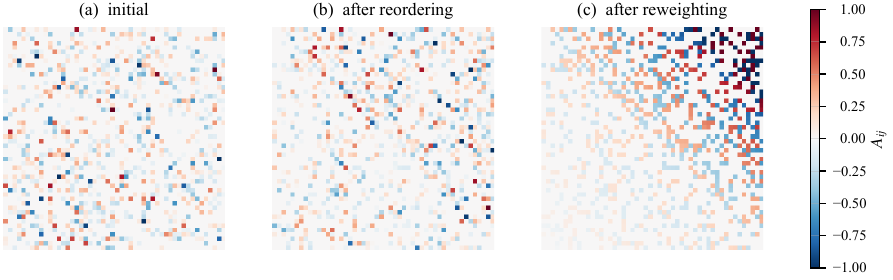}
  \caption{Illustrative example of DAGGER using a smaller network of $n=50$. (a) Heat map of edge weights at initialization. (b) Weights after reordering. (c) Weights after DAGGER reweighting.}
\end{figure}

\section{Method}

\subsection{Approach}

Given the constrained design problem of turning a stable signed sparse matrix $A^0$ into a highly amplifying matrix $A$ that respects the sign/sparsity/diagonal constraints exactly, we seek a method that runs without an iterative inner loop and offers a single tunable knob that traces the trade-off between strict preservation of initial magnitude statistics and aggressive amplification. 

Concretely, given $A^0$ with $A^0_{ii} = -1$, sign pattern $\sigma_{ij} = \mathrm{sign}(A^0_{ij})$, and edge set $\mathcal{E} = \{(i,j) : A^0_{ij} \neq 0\}$, we will look for matrices $A$ that maximize the peak transient response $\mathcal{G}(A) = \sup_{t \geq 0} \lVert e^{At} \rVert_2$ subject to those constraints, a stability margin $\alpha(A) \leq -\rho$, and an optional Wasserstein-2 budget $\mathcal{W}_2(\hat{F}_m, \hat{F}_0) \leq \delta$ on how far the new magnitude distribution $\hat{F}_m$ may drift from the old one $\hat{F}_0$. Setting $\delta = 0$ forces the magnitude multiset of $A$ to be exactly a permutation of the multiset of $A^0$ -- sign, sparsity, and the entire empirical magnitude distribution preserved -- so the only freedom left is which edge gets which magnitude.

Our approach rests on three ideas.  First, transient amplification grows (polynomially or faster) with the length of the longest forward path in the connectivity, where ``forward'' is meant relative to some hierarchical ordering of the units. So pushing connectivity towards a directed-acyclic-graph (DAG) shape is the dominant lever. Second, once we have a per-edge score that measures how amplification-favourable each edge is, the sorted matching -- pair the $k$-th largest magnitude with the $k$-th highest-scoring edge -- provably maximizes any score-weighted magnitude sum among all rearrangements of the multiset onto the edge set (known as the rearrangement inequality \citep{hardy1952}). Third, the natural amplification score for an edge in our setting is its forwardness under the most-DAG-like ordering of the nodes.

Putting these together gives the following procedure: Reorder the nodes to be as DAG-like as possible (Step 1). Score every edge by its forwardness in that ordering, with a small tie-breaker that demotes edges in tight feedback loops (Steps 2–3). Optionally tilt the magnitude multiset to redistribute mass from small magnitudes to large ones (Step 4). Sort-match the (tilted) magnitudes to the edges by score (Step 5). Finally, rescale the off-diagonal block by a single positive constant to land exactly at the stability margin (Step 6). The output respects the sign pattern, sparsity pattern, and diagonal of $A^0$ by construction, and at $\delta = 0$ preserves the magnitude multiset of $A^0$ up to a global scalar.

The remaining subsections give the details and a brief gloss on each step. We describe the strict variant ($\beta = 0$, exact multiset preservation) first; the budgeted version differs only in Step 4.

\subsection{DAG-favouring node ordering}

We need an ordering $\pi$ of the nodes that minimizes the backward-edge weight -- the total magnitude on edges that point against the ordering. Formally we want $\pi$ minimizing $\sum_{(i,j) \in \mathcal{E}, \pi(i) > \pi(j)} |A^0_{ij}|$. This is the weighted minimum-feedback-arc-set problem, which is NP-hard, but a classical greedy heuristic \citep{eades1993} runs in $O(n^2)$ and works well in practice. The procedure is intuitive: at each step look at the residual graph, and if any node has substantially more outgoing weight than incoming, peel it off the front of the ordering as a ``source''; if any node has substantially more incoming than outgoing, peel it off the back as a ``sink''; iterate until the graph is empty. We use this throughout the paper. 

\subsection{Per-edge amplification score}

Given the ordering $\pi$, an edge $(i, j)$ has forwardness $\pi(j) - \pi(i)$: positive if the edge steps downstream, negative if it steps upstream, and large in magnitude if it spans many positions. Forward edges with large forwardness lie on long forward paths and are therefore the structurally amplification-favourable edges. We use $s^{\mathrm{fwd}}_{ij} = \pi(j) - \pi(i)$ as the dominant term in the per-edge score.

We add a small cycle-harm tie-breaker: edges that lie inside tight feedback loops are particularly damaging to amplification because short-cycle dynamics resonate against the spectrum, so for two edges of the same forwardness we should prefer the one that is in a longer cycle (or no cycle at all). Let $L_e$ be the length of the shortest directed cycle through edge $e$, with $L_e = \infty$ if no cycle through $e$ exists. The combined per-edge score is

$$
s_e \;=\; (\pi(j) - \pi(i)) \;-\; \lambda \cdot \frac{1[L_e < \infty]}{L_e},
$$

with $\lambda \geq 0$ a small hyperparameter (we use $\lambda = 0.1$). Computing $L_e$ for all edges takes $O(n |\mathcal{E}|)$ via breadth-first search from each node. The cycle term is structurally dominated by forwardness on densely-cyclic graphs, where the algorithm is essentially insensitive to $\lambda$; on sparser graphs with substantial acyclic substructure the cycle term provides a modest additional benefit.

\subsection{Tilted sort-matching}

We now have everything needed for the assignment step: an edge set $\mathcal{E}$, a per-edge score $\{s_e\}$, and a magnitude multiset $\mathcal{M}_0 = \{|A^0_{ij}|\}$. The sorted matching pairs the largest magnitude with the highest-scoring edge, the second-largest with the second-highest, and so on. By the rearrangement inequality this is the unique maximizer of the score-weighted magnitude sum $\sum_e s_e \, m_{\Pi(e)}$ over all multiset permutations $\Pi$. With the forwardness-based score we have chosen, the sorted matching has the structural effect we want: the largest magnitudes land on edges with the largest rank gaps, which lie on the longest forward paths.

When $\delta = 0$ this is the entire algorithm (modulo the rescale). When $\delta > 0$ we are allowed to tilt the magnitude multiset before the match. We use a one-parameter exponential tilt of the sorted magnitudes,

$$
m'_{(k)} \;=\; m^0_{(k)} \cdot \exp(\beta \cdot u_k), \qquad u_k = \tfrac{k}{K-1} - \tfrac12 \;\in\; [-\tfrac12, +\tfrac12],
$$

with a single hyperparameter $\beta \geq 0$. The 50th-percentile magnitude is preserved ($u = 0$), with magnitudes above the median amplified and below compressed. The amount of mass redistribution is controlled smoothly by $\beta$, and the resulting Wasserstein-2 distance to the original distribution is available in closed form,

$$
\mathcal{W}_2(\hat{F}_m, \hat{F}_0)^2 \;=\; \frac{1}{K} \sum_k \bigl(m^0_{(k)} \cdot (e^{\beta u_k} - 1)\bigr)^2,
$$

so $\beta$ can be tuned to any target $\delta$ by a single scalar root-find. Equivalently, $\beta$ is the additive shift in the log-ratio of the largest-to-smallest magnitude: at $\beta = 0$ the multiset is preserved exactly; at $\beta = k$ the dynamic range $m_{\max} / m_{\min}$ of the sorted multiset is multiplied by $e^k$. So $\beta$ is interpretable as a ``log-spread'' knob: $\beta = 1$ corresponds to a $\sim e \approx 2.7\times$ expansion of dynamic range, $\beta = 3$ to $\sim 20\times$, $\beta = 5$ to $\sim 148\times$.

We chose this exponential-tilt family because it has a clean closed-form $\mathcal{W}_2$, likely admits asymptotic guarantees (the exploration of which we leave for future work), and gives a single interpretable knob; richer tilt families (polynomial, RBF, hinge) achieve more amplification per unit $\mathcal{W}_2$ at small budgets, and we discuss them below.

\subsection{Stability rescaling}

After the assignment in Step 5, the constructed matrix typically has $\alpha(A) \neq -\rho$ — usually too stable, since the rearrangement has clustered the spectrum near $-1$. We close the gap with a single uniform rescaling of the off-diagonal block:
$$
A \;=\; c \cdot N \;+\; \mathrm{diag}(-1, \dots, -1), \qquad c \;=\; \min\bigl\{c' > 0 \;:\; \alpha(c' N + \mathrm{diag}(-1)) \leq -\rho\bigr\},
$$

found by bisection on $c$. The bisection converges in $O(\log(1/\varepsilon))$ calls to a standard eigenvalue routine for the desired tolerance $\varepsilon$; in practice this is around 30 evaluations. The shape of the magnitude distribution is preserved (every off-diagonal entry is multiplied by the same $c$), so the $\mathcal{W}_2$ distance from $\hat{F}_0$ is at most $|c - 1| \cdot \lVert m^0 \rVert_2$ on top of any pre-rescale tilt distance.

The full cost of the algorithm is $O(n^3 \log(1/\varepsilon))$, dominated by the stability bisection (typically $\sim 30$ eigvals at the tolerances we use). This is in sharp contrast to gradient-based methods (SOC, Adam on the numerical abscissa $\omega(A) = \lambda_{\max}\!\bigl(\tfrac12(A + A^\top)\bigr)$, a smooth surrogate for non-normal transient growth), which require hundreds to thousands of eigenvalue evaluations in an inner loop. At $\beta = 0$, DAGGER's output preserves the magnitude multiset of $A^0$ exactly up to a global scalar (the rescale factor $c$). 

\subsection{Optional refinement at strict preservation}

When $\delta = 0$ is required and we want to push as far as possible within the strict variant, a brief swap-refinement step yields a useful additional gain. We compute the leading eigenvector $v$ of $(A + A^\top)/2$, form the per-edge $\omega$-gradient $g_e = \sigma_e \, v_i \, v_j$, and re-sort-match the (unchanged) multiset against $\{g_e\}$. We accept the new assignment if it preserves stability, otherwise binary-search for the largest prefix of swaps that does. Running this from several random initializations and selecting by peak amplification rather than $\omega$ consistently improves on a single sort-matched start by a substantial factor.  This step is optional: even without it the strict variant matches or exceeds existing methods at multiset preservation.

\section{Experiments}

We evaluate DAGGER against three competitor methods plus a theoretical reference on five classes of experiments: a Pareto-frontier comparison on synthetic substrates, a wall-clock speed benchmark across substrate sizes, a pseudospectral diagnostic that explains why DAGGER amplifies, a downstream signal-detection task, and structural diagnostics.

\subsection{Setup}

Synthetic substrates are random sparse signed matrices at density $0.18$, off-diagonal magnitudes drawn from $\text{lognormal}(\mu = -0.5, \sigma = 0.8)$ unless noted, signs uniformly random, diagonal fixed at $-1$. The lognormal choice with $\sigma$ in the $0.8$–$1.5$ range is consistent with measured synaptic-strength distributions in mammalian cortex \citep{song2005, buzsaki2014}. Each substrate is bisection-rescaled at sample time to a uniform spectral abscissa $\alpha = -0.03$. The Pareto comparison uses $n = 200$ across 10 seeds; the speed sweep uses $n \in \{60, 80, 120, 200, 300\}$ across 10 seeds; the pseudospectral diagnostic and detection task  use $n = 80$. We verify in Appendix B that DAGGER's qualitative advantages -- gradient-free single-pass construction, single-knob Pareto, asymptotic unboundedness, and a peak-amplification advantage over SOC that grows with the dispersion of the starting magnitude distribution -- are robust across the input magnitude distribution (lognormal at three $\sigma$ values, uniform, exponential).

Methods compared:
\begin{itemize}
\item L0 — uniform off-diagonal rescaling to the target stability margin, no other change.
\item SOC — Stability-Optimised Circuit \cite{hennequin2014, stroud2018}. The original method optimizes a smoothed spectral abscissa under Dale's-law sign constraints and element-wise magnitude bounds drawn from biological data. For an apples-to-apples comparison against DAGGER's Wasserstein-2 budget on the aggregate magnitude distribution, we substitute periodic Frobenius-norm preservation, which constrains total magnitude similarly. We use a corrected greedy local-search variant (5 candidate perturbations per step, 300 steps).
\item GRAD-Adam — gradient ascent on $\omega(A) = \lambda_\text{max}((A + A^\top)/2)$ via Adam, sign-clipped, Frobenius-projected, 5-restart multistart, 1000 steps per restart.
\item DAGGER — our method, with $\beta \in \{0, 0.5, 1, 1.5, 2, 3, 5\}$ in the Pareto sweep (Figure 2) and an extended sweep up to $\beta = 40$ in Appendix C; $\beta = 0$ is the strict multiset-preserving variant. The detection task (§4.5) shows three representative $\beta$ values: 0, 3, 6.
\end{itemize}

All methods are post-rescaled to $\alpha = -0.03$ to ensure equal-stability comparison.

Note on the SOC adaptation: the SOC variant we compare against is an adaptation of the classical algorithm to our constrained-design setting. The original SOC procedure samples a random unstable matrix and optimizes toward stability, using the chaotic initialization as an implicit reservoir of non-normality that survives the gradient flow. Our problem cannot do this: $A^0$'s sign, sparsity, diagonal, and (at $\beta = 0$) magnitude multiset must all be preserved exactly, and there is no constrained "unstable initialization" of $A^0$ available. Our adapted SOC therefore initializes at $A^0$, runs gradient descent on $\alpha$ (driving $\alpha$ further negative, with Frobenius-norm preservation), and post-rescales the off-diagonal to recover the common target $\alpha = -0.03$. This is the minimum modification required to apply the SOC algorithmic core under hard substrate constraints; the original procedure does not preserve $A^0$ as a substrate.

\subsection{Pareto frontier and competitor comparison}

\begin{figure}
  \centering
  \includegraphics[width=0.99\textwidth]{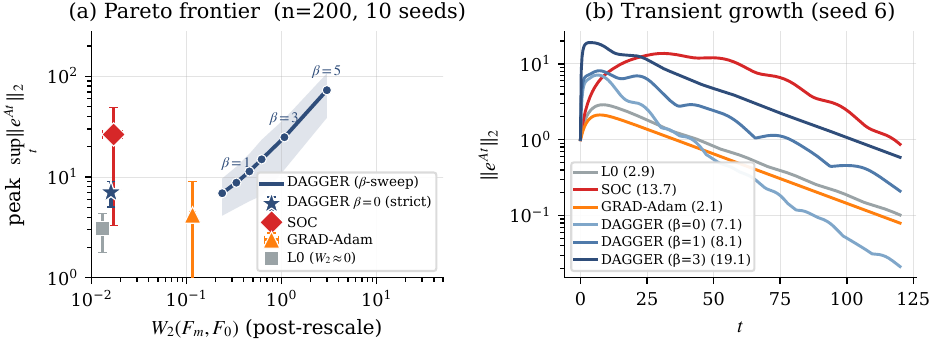}
  \caption{Peak amplification vs achieved Wasserstein-2 budget, post-rescale, on $n = 200$ synthetic substrates (10 seeds, mean $\pm$ std). (a) log-log Pareto frontier. (b) Transient growth curves on a single representative seed.}
\end{figure}

DAGGER's $\beta$-sweep traces a continuous, monotone Pareto curve from peak $\approx 7$ at $W_2 \approx 0.09$ ($\beta = 0$, strict multiset) to peak $\approx 73$ at $W_2 \approx 3$ ($\beta = 5$). SOC sits at $W_2 \approx 0.02$ with mean peak $\approx 26$ and large seed-to-seed variance (std $\approx 23$, coefficient of variation $\approx 0.9$). DAGGER $\beta = 5$ exceeds SOC's mean peak by $\sim 3\times$ on average and is also far more reliable per seed (std $\approx 34$ on a mean of $73$, CV $\approx 0.5$). GRAD-Adam is much weaker than either (peak $\approx 4$). As $\beta$ grows, DAGGER sweeps several orders of magnitude of peak amplification with smoothly increasing $W_2$, while SOC and GRAD-Adam remain at single fixed points with no comparable budget knob.

Asymptotic behaviour: peak amplification grows essentially without bound in $\beta$; on a wide-range sweep ($\beta \in [0, 40]$, Appendix C) we observe $\log(\text{peak})$ growing approximately linearly with empirical slope $\approx 0.58$, reaching peak $\approx 10^{10}$ at $\beta = 40$ before float64 overflow at higher $\beta$. SOC and GRAD-Adam have no analogous knob; their amplification is bounded by their convergence basin and cannot be pushed further with additional compute.

\subsection{Wall-clock speed and complexity}

Figure 3(a) shows wall-clock time vs $n$ on log-log axes. DAGGER scales as $O(n^3 \log(1/\varepsilon))$; SOC and GRAD-Adam scale as $O(n^3 \cdot T)$ with $T = 1500$ (SOC: 300 steps × 5 candidates per step) and $T = 200$ (GRAD-Adam) respectively in this configuration. The cost in eigendecompositions per construction is shown in panel (b): DAGGER $\approx 45$ (30 stability bisection + 15 swap-opt), SOC $= 1500$, GRAD-Adam $= 200$.

\begin{figure}
  \centering
  \includegraphics[width=0.99\textwidth]{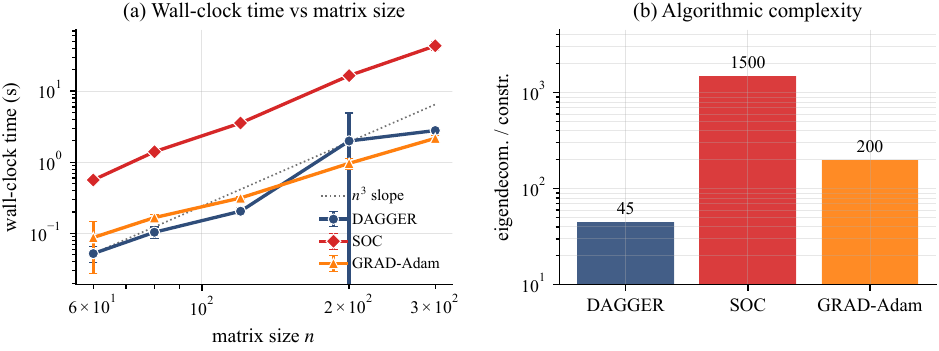}
  \caption{Speed comparison, swept across $n \in \{60, 80, 120, 200, 300\}$ over 10 seeds. (a) Wall-clock time vs $n$ on log-log axes. (b) Eigendecompositions per construction.}
\end{figure}

At $n = 300$ DAGGER takes $\approx 3$ s per construction against SOC's $\approx 16$ s and GRAD-Adam's $\approx 2$ s. The SOC step count used here (300) is conservative relative to the canonical $\sim 1{,}000$ steps reported by \citep{hennequin2014}; the full-step SOC would be $\approx 5\times$ slower again. The wall-clock advantage scales with $n$ in the same way as the algorithmic complexity -- DAGGER is gated by a logarithmic number of eigendecompositions while SOC and GRAD-Adam carry a multiplicative constant, so the absolute time gap grows with $n$ even though the ratio is roughly constant.

\subsection{Pseudospectral diagnostic: why DAGGER amplifies}

The $\varepsilon$-pseudospectrum $\sigma_\varepsilon(A) = \{z : \sigma_\text{min}(zI - A) < \varepsilon\}$ encodes the spatial extent over which the resolvent norm exceeds $1/\varepsilon$. For non-normal matrices this set extends into the right half-plane even when all eigenvalues sit in the left half-plane; the rightmost point of $\sigma_\varepsilon(A)$, called the pseudospectral abscissa $\alpha_\varepsilon$, directly bounds the peak transient amplification \citep{trefethen2005}. Methods that produce more peak amplification at matched $\alpha(A)$ have pseudospectra that extend further to the right.

\begin{figure}
  \centering
  \includegraphics[width=0.99\textwidth]{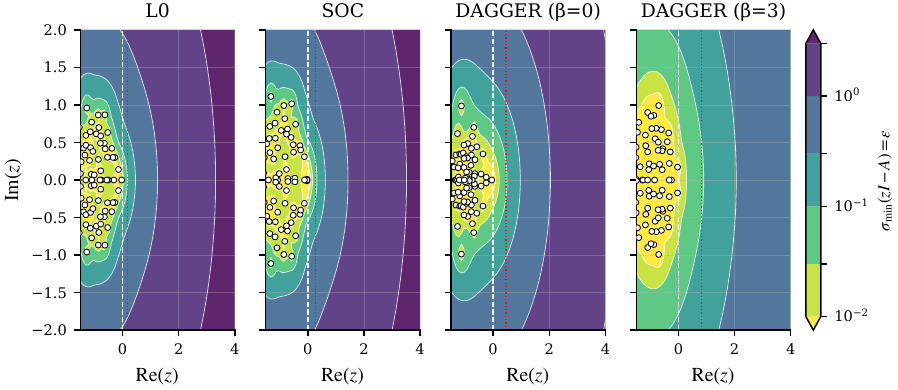}
  \caption{$\epsilon$-pseudospectra of L0, SOC, DAGGER ($\beta=0$), and DAGGER ($\beta=3$) on a representative matrix (n=80, seed 2, post-rescale $\alpha = -0.03$). Filled contours show level sets of $\sigma_\text{min}(zI - A)$ from $10^{-2}$ to $3$; eigenvalues are marked as white dots; the white dashed line is the imaginary axis (stability boundary).}
\end{figure}

Figure 4 makes the comparison directly visible. The pseudospectral abscissae $\alpha_{\varepsilon}$ at $\varepsilon = 0.1$ for this matrix are: L0 = $0.18$, SOC = $0.27$, DAGGER ($\beta = 0$) = $0.46$, and DAGGER ($\beta = 3$) = $0.83$. L0 and SOC have pseudospectra that hug their eigenvalues: modest non-normality, $\alpha_{0.1}$ near the spectral abscissa. DAGGER ($\beta = 0$) more than doubles SOC's pseudospectral reach despite preserving a comparable magnitude multiset, and DAGGER ($\beta = 3$) bulges substantially further into the right half-plane with $\alpha_{0.1}$ approaching unity. This is the structural signature of DAGGER's amplification advantage: not eigenvalue movement (which is forbidden by the rescaling step), but pseudospectral bulge driven by the rearrangement of magnitudes onto forward-scoring edges.

\subsection{Downstream signal-detection task}
We adapt the standard non-normal optimal-input construction \citep{trefethen2005,hennequin2014} to a binary signal-detection paradigm in the spirit of \citep{murphy2009, bondanelli2020}: each method's optimal-amplification direction $u^\star$ is taken as the leading right singular vector of $e^{A t^\star}$, and on each trial the network's input is either an aligned signal or pure isotropic noise; the network's transient response is read out at its peak. We sweep signal amplitude $a \in [0.0003, 1.0]$ and report ROC AUC.

\begin{figure}
  \centering
  \includegraphics[width=0.99\textwidth]{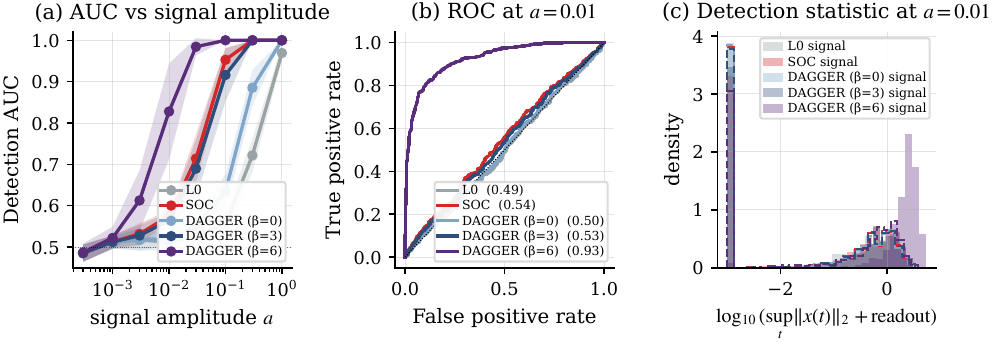}
  \caption{Signal-detection task, showing (a) AUC vs signal amplitude, log $a$-axis; (b) ROC curves at $a=0.01$ for a single representative seed; legend values are that seed's AUC; and (c) Detection-statistic histograms (signal vs null) at the same amplitude.}
\end{figure}

DAGGER at $\beta=6$ dominates detection, reaching AUC = 1 at $a = 0.03$ -- an order of magnitude weaker signal than any competitor. At $a = 0.01$, $\beta=6$ reaches AUC $\approx 0.87$ vs. SOC's $\approx 0.62$, while DAGGER at lower $\beta$ remains near chance (AUC $\approx 0.50-0.55$), comparable to the L0 and SOC baselines at this amplitude. The advantage emerges at high $\beta$, where the tilt more aggressively reshapes the substrate, and scales with peak amplification -- confirming that the algorithmic advantage transfers to functional readout.

\subsection{Structural diagnostics: why DAGGER amplifies}

\begin{table}
  \caption{Structural metrics per method on a representative matrix ($n = 200$, seed $7$, $\alpha = -0.03$).}
  \label{beta}
  \centering
  \begin{tabular}{lrrrr}
    \toprule
   Method & Peak $\mathcal{G}^\star$ & Eff\_rank($A$) & Eff\_rank($e^{A t^\star}$) & Bwd-fraction \\
    \midrule

L0 & 1.71 & 137.20 & 15.27 & 0.376 \\
SOC & 10.07 & 130.07 & 3.35 & 0.374 \\
DAGGER ($\beta = 0$) & 3.67 & 77.80 & 17.44 & 0.176 \\
DAGGER ($\beta = 1$) & 5.73 & 56.09 & 18.13 & 0.111 \\
DAGGER ($\beta = 3$) & 15.82 & 36.30 & 16.68 & 0.041 \\
DAGGER ($\beta = 6$) & 87.59 & 27.36 & 14.06 & 0.008 \\
    
    \bottomrule
  \end{tabular}
\end{table}

Table 1 shows two findings that discriminate the methods cleanly, on a representative sample matrix. First, backward-edge fraction drops monotonically with $\beta$: $0.176$ ($\beta = 0$) $\to 0.008$ ($\beta = 6$), a reduction of $22\times$. SOC's backward fraction is essentially identical to L0 (0.374 vs 0.376) and more than 40$\times$ larger than DAGGER at $\beta = 6$, confirming that SOC modifies magnitudes without reshaping topology. DAGGER's amplification gain is therefore a topology gain in the original signed sparse basis, even though the sparsity pattern itself is preserved exactly. Second, structural effective rank drops from 137.20 (L0, near full rank) to 27.36 (DAGGER $\beta = 6$, structurally low-rank), reflecting concentration of magnitude on high-scoring edges; transient effective rank stays moderate ($\sim$14–18 for DAGGER / L0 vs 3.4 for SOC), indicating that DAGGER networks are multi-mode amplifiers with several long forward chains contributing to several near-peak singular values of $e^{At^\star}$ simultaneously, while SOC concentrates almost all of its amplification onto a single mode. This multi-mode property is a feature: a network amplifying several input directions simultaneously is more useful for general signal processing than one locked to a single mode.

\section{Discussion}
Constraint structure can replace optimization when the geometry is rich enough. DAGGER is the right tool when (a) the substrate connectivity must be preserved exactly -- biological connectomes, Dale's-law-respecting models, initialized RNNs with prescribed sparsity -- and (b) the user wants a fast, single-pass construction with an interpretable trade-off knob. The gradient-free design eliminates the inner loop of SOC, giving 30-100$\times$ fewer eigendecompositions per construction (and roughly 5$\times$ wall-clock speedup against SOC at $n = 300$), and the single scalar $\beta$ traces the entire (W$_2$, $\mathcal{G}$) Pareto front from strict multiset preservation to asymptotically unbounded amplification. Crucially, the amplification gain comes from rearrangement of magnitudes onto forward-scoring edges in the original signed sparse basis -- not from changing connectivity -- which is what is needed when working with biological substrates.

This work provides structural insight as an alternative to optimization. DAGGER's advantage over SOC is not just computational speed. The two methods address the same constrained problem, but DAGGER replaces SOC's gradient loop with a closed-form analytic assignment via the rearrangement inequality. This is possible because the constraint structure itself -- fixed sign, sparsity, diagonal, and (at $\beta = 0$) magnitude multiset -- restricts the search space to multiset permutations onto edges, where the rearrangement inequality applies. In unconstrained or loosely-constrained settings, gradient methods are the right tool; under hard structural constraints, the geometry of the constraint set can enable analytic optima that gradient descent only approximates. We see DAGGER as one instance of a more general pattern: when the constraint structure is rich enough, structural insight can replace optimization. 

Three clear limitations of this work from an algorithm standpoint: 
\begin{enumerate}
\item Topology held fixed: DAGGER holds the sparsity pattern fixed by design; an unconstrained nilpotent ceiling (the limiting case described in \citep{goldman2009}) sits roughly $10^9$ above any constrained method, indicating that some applications could productively trade off topology preservation for amplification.  
\item Exponential tilt is not optimal: the exponential tilt was chosen for its closed-form W$_2$ distance and asymptotic guarantee, but is empirically not the optimal tilt under our constraints.
\item Greedy ordering is heuristic: spectral and PageRank-style alternatives substitute trivially with mild empirical effect, and we do not claim our heuristic is globally optimal in any precise sense. 
\end{enumerate}
A further possible limitation: despite its computational advantage, DAGGER's multi-mode amplifying structure may not be the best fit when a single-mode amplifier is preferred.

There are numerous interesting future directions worth exploring. These include a deeper study of the differences between SOC structure and DAGGER structure; an application of DAGGER to large biological connectomes, which are computationally out of reach for SOC; and an investigation of whether using DAGGER as initialization for nnRNNs \citep{kerg2019} or other Schur-parameterized architectures improves training dynamics.
\newpage

\bibliographystyle{abbrvnat} \bibliography{references}

@article{kerg2019,
  title={Non-normal recurrent neural network (nnrnn): learning long time dependencies while improving expressivity with transient dynamics},
  author={Kerg, Giancarlo and Goyette, Kyle and Puelma Touzel, Maximilian and Gidel, Gauthier and Vorontsov, Eugene and Bengio, Yoshua and Lajoie, Guillaume},
  journal={Advances in neural information processing systems},
  volume={32},
  year={2019}
}

@article{christodoulou2022,
  title={Regimes and mechanisms of transient amplification in abstract and biological neural networks},
  author={Christodoulou, Georgia and Vogels, Tim P and Agnes, Everton J},
  journal={PLoS computational biology},
  volume={18},
  number={8},
  pages={e1010365},
  year={2022},
  publisher={Public Library of Science San Francisco, CA USA}
}

@article{stroud2018,
  title={Motor primitives in space and time via targeted gain modulation in cortical networks},
  author={Stroud, Jake P and Porter, Mason A and Hennequin, Guillaume and Vogels, Tim P},
  journal={Nature neuroscience},
  volume={21},
  number={12},
  pages={1774--1783},
  year={2018},
  publisher={Nature Publishing Group US New York}
}

@article{hennequin2014,
  title={Optimal Control of Transient Dynamics in Balanced Networks Supports Generation of Complex Movements},
  author={Hennequin, Guillaume and Vogels, Tim P and Gerstner, Wulfram},
  journal={Neuron},
  volume={82},
  pages={1394--1406},
  year={2014},
}

@article{murphy2009,
title = {Balanced Amplification: A New Mechanism of Selective Amplification of Neural Activity Patterns},
journal = {Neuron},
volume = {61},
number = {4},
pages = {635-648},
year = {2009},
author = {Brendan K. Murphy and Kenneth D. Miller}
}

@article{orhan2019,
  title={Improved memory in recurrent neural networks with sequential non-normal dynamics},
  author={Orhan, A Emin and Pitkow, Xaq},
  journal={arXiv preprint arXiv:1905.13715},
  year={2019}
}

@article{Goldman2009,
  title={Memory without feedback in a neural network},
  author={Goldman, MS},
  journal={Neuron},
  volume={61},
  pages={621--634},
  year={2009},
}

@article{bondanelli2020,
  title={Coding with transient trajectories in recurrent neural networks},
  author={Bondanelli, Giulio and Ostojic, Srdjan},
  journal={PLoS computational biology},
  volume={16},
  number={2},
  pages={e1007655},
  year={2020},
  publisher={Public Library of Science San Francisco, CA USA}
}

@article{bondanelli2021,
  title={Network dynamics underlying OFF responses in the auditory cortex},
  author={Bondanelli, Giulio and Deneux, Thomas and Bathellier, Brice and Ostojic, Srdjan},
  journal={Elife},
  volume={10},
  pages={e53151},
  year={2021},
  publisher={eLife Sciences Publications, Ltd}
}

@inproceedings{arjovsky2016,
  title={Unitary evolution recurrent neural networks},
  author={Arjovsky, Martin and Shah, Amar and Bengio, Yoshua},
  booktitle={International conference on machine learning},
  pages={1120--1128},
  year={2016},
  organization={PMLR}
}

@inproceedings{mhammedi2017,
  title={Efficient orthogonal parametrisation of recurrent neural networks using householder reflections},
  author={Mhammedi, Zakaria and Hellicar, Andrew and Rahman, Ashfaqur and Bailey, James},
  booktitle={International Conference on Machine Learning},
  pages={2401--2409},
  year={2017},
  organization={PMLR}
}

@inproceedings{lezcano2019,
  title={Cheap orthogonal constraints in neural networks: A simple parametrization of the orthogonal and unitary group},
  author={Lezcano-Casado, Mario and Mart{\'\i}nez-Rubio, David},
  booktitle={International Conference on Machine Learning},
  pages={3794--3803},
  year={2019},
  organization={PMLR}
}

@book{trefethen2005,
  title={Spectra and pseudospectra: the behavior of nonnormal matrices and operators},
  author={Trefethen, Lloyd N and Embree, Mark},
  year={2020},
  publisher={Princeton university press}
}

@article{lewis1996,
  title={Eigenvalue optimization},
  author={Lewis, Adrian S and Overton, Michael L},
  journal={Acta numerica},
  volume={5},
  pages={149--190},
  year={1996},
  publisher={Cambridge University Press}
}

@book{hardy1952,
  title={Inequalities},
  author={Hardy, Godfrey Harold and Littlewood, John Edensor and P{\'o}lya, George},
  year={1952},
  publisher={Cambridge university press}
}

@article{vanbiervliet2009,
  title={The smoothed spectral abscissa for robust stability optimization},
  author={Vanbiervliet, Joris and Vandereycken, Bart and Michiels, Wim and Vandewalle, Stefan and Diehl, Moritz},
  journal={SIAM Journal on Optimization},
  volume={20},
  number={1},
  pages={156--171},
  year={2009},
  publisher={SIAM}
}

@article{eades1993,
  title={A fast and effective heuristic for the feedback arc set problem},
  author={Eades, Peter and Lin, Xuemin and Smyth, William F},
  journal={Information processing letters},
  volume={47},
  number={6},
  pages={319--323},
  year={1993},
  publisher={Elsevier}
}

@article{song2005,
  title={Highly nonrandom features of synaptic connectivity in local cortical circuits},
  author={Song, Sen and Sj{\"o}str{\"o}m, Per Jesper and Reigl, Markus and Nelson, Sacha and Chklovskii, Dmitri B},
  journal={PLoS biology},
  volume={3},
  number={3},
  pages={e68},
  year={2005},
  publisher={Public Library of Science San Francisco, USA}
}

@article{buzsaki2014,
  title={The log-dynamic brain: how skewed distributions affect network operations},
  author={Buzs{\'a}ki, Gy{\"o}rgy and Mizuseki, Kenji},
  journal={Nature reviews neuroscience},
  volume={15},
  number={4},
  pages={264--278},
  year={2014},
  publisher={Nature Publishing Group UK London}
}

\newpage

\appendix

\renewcommand{\thefigure}{\thesection.\arabic{figure}} 
\setcounter{figure}{0} 

\section{Computational Resources}
Experiments were conducted on a single laptop (Apple M4 Pro 24GB RAM, 16‑Core GPU). Each matrix construction completed in seconds to minutes; the full experimental sweeps reported in the paper run in well under an hour on this hardware. Code was implemented in Python 3.13.

\section{Cross-distribution robustness}

We test DAGGER's qualitative advantages across five starting magnitude distributions: lognormal at $\sigma \in \{0.4, 0.8, 1.5\}$, uniform on $[0.1, 1.1]$, and exponential (rate 1). For each distribution we sample $n = 200$ matrices at density 0.18 and target $\alpha = -0.03$ over 5 seeds, and run DAGGER ($\beta$-sweep), SOC, and GRAD-Adam.

\begin{figure}
  \centering
  \includegraphics[width=0.99\textwidth]{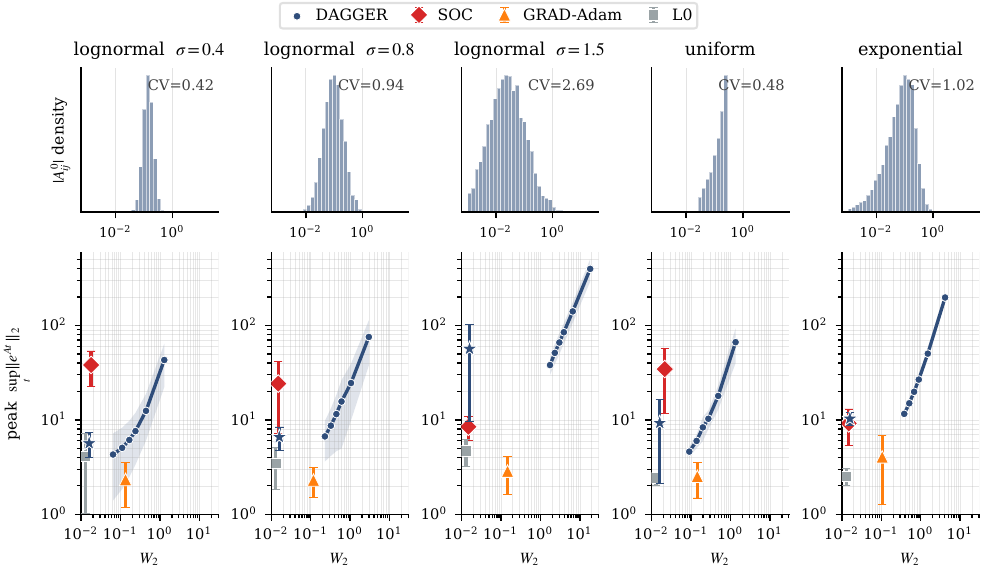}
  \caption{Pareto frontiers per starting distribution. Top row: histograms of off-diagonal magnitudes (with coefficient of variation). Bottom row: peak amplification vs $W_2$ for each method, mean ± std across 5 seeds.}
\end{figure}

Three observations: (1) DAGGER's advantage scales with input dispersion (CV). At low CV (0.42, 0.48), SOC mean $\approx$ DAGGER $\beta=5$; at moderate CV (0.94, 1.02), DAGGER $\beta=5$ dominates by 3–20×; at high CV (2.69, lognormal-1.5), DAGGER $\beta=5$ mean is $\sim 50\times$ SOC mean. Cortical synaptic-strength data lives at CV $\geq 1$ \citep{song2005,buzsaki2014}, the regime where DAGGER's advantage is clearest. (2) SOC's variance is large in the narrower-distribution regimes (lognormal-0.4 and -0.8, CVs of 0.42 and 0.94) and small in heavy-tailed regimes -- SOC's gradient flow is highly sensitive to initialization when the magnitude landscape is uniform, and largely insensitive when there is a clear ordering of important edges in the input. DAGGER's variance is small across all regimes. (3) GRAD-Adam is consistently weak, with peak $\sim 2$ -- $4$ across all distributions.

These observations are consistent with the structural intuition that DAGGER's tilt has more to push when the original spread is wide; the $W_2$-budgeted exponential tilt is most effective on multisets that already have headroom in their dynamic range.

\section{Asymptotic amplification}

Peak amplification grows essentially without bound in $\beta$. Across three seeds at $n = 80$ we observe $\log(\text{peak})$ growing approximately linearly with empirical slope $\approx 0.58$, reaching peak $\approx 10^{10}$ at $\beta = 40$ before float64 overflow at higher $\beta$ values (the rescale factor $c$ becomes too small for double precision). The theoretical lower bound $\beta/2$ from the rearrangement inequality applied to a generic sort-matched multiset is recovered up to a constant; the gap $0.58 - 0.50 = 0.08$ measures the additional advantage from DAGGER's structural alignment with the longest forward chain. This is what "asymptotic unboundedness" means in the main text — the $\beta$ knob is not bounded by any algorithmic feature internal to DAGGER.

\begin{figure}
  \centering
  \includegraphics[width=0.6\textwidth]{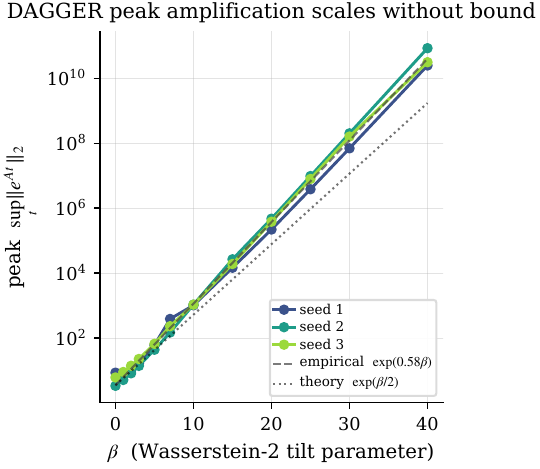}
  \caption{$\log(\text{peak})$ vs $\beta$ on a wide-range sweep ($\beta \in [0, 40]$) at $n = 80$, density 0.18, three seeds. Solid line: empirical fit, $\log(\text{peak}) \approx 0.58\,\beta + c$, fitted on the upper 60\% of the $\beta$ range (asymptotic regime). Dashed line: theoretical lower bound $\log(\text{peak}) \geq \beta/2$, anchored to the empirical intercept.}
\end{figure}

\end{document}